# Synthetic Counteradaptation

## A Principle of Human–AI Co-evolution

Ivar Frisch[1], Jackie Kay[2], Philip Moreira Tomei[3]

[1]*Spectral Circuits Research* [2]*Independent Researcher* [3]*AI Objectives Institute*

**Abstract**

In this paper, we introduce the concept of *synthetic counteradaptation*, a process where human and AI systems co-evolve by adapting to each other's strategies and behaviors. Synthetic counteradaptation occurs when AI systems develop novel strategies or social protocols, prompting humans to extract insights and adapt their own behaviors in response, leading to the emergence of new agent interaction dynamics. To illustrate these dynamics, we analyze examples from various contexts, including the game of Go, mixed-motive social interactions, and geopolitical simulations. By exploring these cases, we demonstrate how synthetic counteradaptation provides a framework for understanding the recursive and co-evolutionary nature of human–AI interactions in multi-agent environments.

---

## 1 Introduction

> *Technology, like biology, does not exist in the absence of evolution. Technology is not artificially replacing life—it is life.*
>
> —Sara Walker [35]

In today's world, technology and (biological) life are often seen as opposites: artificial, man-made objects versus naturally-occurring organisms. However, the work of philosophers and cyberneticians such as Gilbert Simondon and Ross Ashby already blurred this distinction between life and technology. They, amongst many mid-20th century systems philosophers, understood technology not as opposed to life, but rather as an extension of evolutionary and adaptive processes [16]. This precedent invites us to consider the potential for treating AIs as adaptive agents who participate autonomously in life's evolutionary patterns. For instance, it has been shown that societies of LLM agents can exhibit emergent multi-agent phenomena such as information diffusion [25] and linguistic alignment [13]—phenomena which emerged in humans over millennia of evolution. However, within the field of AI, the co-evolution between machinic and human agents is under-specified—and the perspective that technology and life are extensions of one another is rarely taken for granted. This leaves a gap of insights into the principles and dynamics of co-evolution and their impact on AI and human development.

This paper explores one principle of human–AI co-evolution, *synthetic counteradaptation*, as a thought experiment, leveraging current debates on synthesis and counteradaptation in order to understand the implications of the rise of AI systems for the future development of life's evolutionary patterns. *Synthetic* refers to the relational interplay between a natural and an artificial agent, where their relational dependencies create a relational meaning of them and, thus, establishes them as a systemic whole within a coevolutionary process. *Counteradaptation* refers to the adjustments made by one organism in response to the adaptations of another. It represents a specific aspect of the broader coevolutionary process, focussing on how one species develops traits that counteract or mitigate the effects of adaptations made by another species. Counteradaptation is a fundamental aspect of co-evolution, illustrating how species interact and influence each other's evolutionary trajectories. The interplay of adaptations and counteradaptations not only shapes the traits of individual species but also drives broader patterns of biodiversity and ecological dynamics [9]. Two dynamics of counteradaptation are relevant for our discussion:

- **Adversariality:** Coevolution often results in evolutionary arms races, where species continuously adapt and counter-adapt in response to each other. This can be seen in predator–prey dynamics, where predators evolve better hunting strategies, while prey develop more effective escape mechanisms.
- **Mutualism:** In mutualistic relationships, counteradaptation can also occur. For example, flowering plants can develop traits that attract pollinators, while pollinators can adapt to become more efficient at extracting nectar. This mutual influence exemplifies co-evolution through counteradaptation [6].

As such, we will propose the process of synthetic counteradaptation which arises within human–AI interaction. First, we will demonstrate how the principle of synthetic counteradaptation operates in the game of Go. Next, we will give a theoretical exposition that elucidates the concepts of "synthesis" and "counteradaptation." Finally, we will present some speculative scenarios on the future of human–AI interaction through the lens of the process of synthetic counteradaptation.

In the game of Go, we contextualise three instances of human–AI interaction. We begin with the historic four game series hosted in 2016, between Lee Sedol, a 9-dan professional ranked 1st worldwide at the time, and AlphaGo, an AI trained to play Go via reinforcement learning and self-play. After winning the first game, AlphaGo stunned the international Go community in the second game at Move 37, an idiosyncratic play that was understood (after post-hoc analysis) to be a significant innovation in the game of Go. In the fourth game, Lee Sedol adopted a risky strategy known as *amashi*. He played the lesser known but equally important Move 78, a move that was as equally unexpected as Move 37, and secured victory. Although Lee Sedol still lost the overall series, Move 78 illustrated his capacity for rapid adaptation to AlphaGo's playing style.

In more recent developments, researchers at FAR AI, an AI safety evaluation company [38], developed an AI tutor that leveraged an adversarial training update in games against AI Go players that outperform humans [36]. This AI tutor was able to extract a specific strategy to exploit the playing style of this latter class of AI players, which the researchers used to teach amateur human players to beat these state-of-the-art AI agents.

In this progression, we first observed the dynamics of synthetic counteradaptation, which can be broken down into three steps. *Mutation* is the first step in evolution, when an initial change occurs in agents or conditions of the environment. Consequently, *adaptation* is when an agent devises or evolves an adaptation to this change. Finally, *counteradaptation* is when another agent adapts to the prior adaptation. To summarize the example of the game of Go in these steps:

1. **Mutation.** Move 37: AlphaGo discovers a novel strategy.
2. **Adaptation.** Move 78: Lee Sedol adapts to AlphaGo's playing style.
3. **Counteradaptation.** Move 349: AI teaches amateurs an exploit to beat AlphaGo-level AI players.

Thus, we have shown how the principle of synthetic counteradaptation operates in the game of Go. However, what does this principle actually mean? How does it relate to wider strands of philosophy, agent interaction and evolutionary theory? In order to elucidate this, the next section will explain the notion of synthetic counteradaptation by drawing on the Hegelian notion of synthesis and evolutionary theory.

## 2 Synthetic Counteradaptation

We articulate the meaning of synthetic counteradaptation by breaking down the term into its two constituent parts.

### 2.1 Synthetic

The meaning of the term "synthetic" is twofold. In the first, we refer to the naive definition, substances that are artificially created or human made such as lab-grown diamonds. In synthetic chemistry, for example, compounds are artificially produced through chemical reactions, mimicking or modifying naturally occurring substances. This includes the creation of pharmaceuticals, plastics, and other industrial chemicals. Similarly, in biology, *synthetic biology* involves the design and construction of new biological parts, systems, or even entire organisms that do not exist in nature, often using genetic engineering techniques to modify DNA for specific purposes. Although this notion of synthetic highlights the idea that something is artificial, it does not yet highlight a relation between the "natural" entity and the "artificial" entity. Neither does it allow for a scaffolding of this relationship between the two entities. Both of which are needed in order to characterize the evolutionary relationship between (human) organisms and machines.

As such, we also refer to the Hegelian notion of "synthesis" (otherwise known as 'Aufhebung'). For Hegel, synthesis refers to the 'final' moment in the process of dialectics where two, seemingly, contradictory entities are reconciled and their logical structure is shown to be interrelated. For example, in the *Science of Logic* [15], Hegel shows that the concepts *Being* and *Nothing* contain one another, because Being considered on its own is so empty, it contains so little, that it is in fact *Nothing*. Conversely, the concept of *Nothing* exists (it *is*), because *Nothing* is something which can be imagined by thought. As such, the two seemingly oppositional concepts are shown to be interrelated and are thus synthesized into a higher-order concept (in this case "*Becoming*") [15].

Although Hegel here talks specifically about concepts, his notion of synthesis can be used to characterize human–AI relations as well. This is mainly due to the fact that Hegel's notion of synthesis has been shown to be applicable to the constitutive relationship of organisms with their environment [39] and to the idea of AI being the evolutionary successor of humanity in terms of perpetuating the same fundamental operation of intelligence [40].

As such, the important takeaway from Hegel's notion of synthesis is that this form of synthesis forms a recursive deepening of the meaning of the constituent entities, by virtue of the negation of the individual terms into a higher-order relation. Both terms are connected via a higher-order relation (*Becoming*) and are thus seen as parts in a larger whole. They now derive their meaning from being parts of a whole rather than being a whole in themselves. However, they are not completely dissolved. A complete resolution is never obtained [41]. *Becoming* needs both the meaning of *Being* and *Nothingness* in order for it to denote a process by which things come into existence or go out of existence.

As such, the moment of synthesis denotes a continuous process of scaffolding. "Aufhebung is the suppression that conserves" [24]. The individual entities form the building blocks for higher-order relations and these higher order relations will themselves also serve as building blocks for further higher-order relations.

Putting the two notions together, we define "synthetic" as the moment in which an artificially created agent is shown to be interrelated with an organic agent. This moment of synthesis forms a recursive deepening of the individual agents, wherein both agents get a meaning in terms of the relation they have to each other.

## 2.2 Counteradaptation

> *The complex adaptations and counteradaptations we see between predators and their prey are testament to their long coexistence and reflect the result of an arms race over evolutionary time.*
>
> —John R. Krebs [20]

*Counteradaptation* occurs when an agent adapts to the prior adaptation of another organism, which is itself a reaction to a mutation in the agent or environment.

In molecular biology, a mutation is a change in the genetic sequence of an organism [11]. Based on this, we use the term mutation in order to denote any genetic change in the agent or environment. As such, our principle of synthetic counteradaptation can be seen as a principle applicable to any scale of agent interaction.

When an organism responds to a mutation in another agent or to a mutation in the environment, this can be defined as *adaptation*. We identify three forms of adaptations: evolutionary adaptation, physiological adaptation, cognitive adaptation. *Evolutionary adaptation* [1] is when genetic traits are selected as solutions to a specific problem, for a particular function or purpose. This evolutionary adaptation is a long process of genetic mutation and selection across generations and over time. As such, it is different from a physiological adaptation. *Physiological adaptation* is an immediate organismal response to a particular stress factor and can happen instantaneously: if you heat up from the sun your nervous system responds by sweating.

Another form of adaptation can also be identified: *cognitive adaptation*. Cognitive adaptation can operate on a larger, global scale between agents. As such, it denotes the development of "evolving tactics, technologies, targets, group dynamics, and other behaviors" [14]. Thus, in cognitive adaptation we find the emergence of more complex phenomena such as mimetic mind modelling. This denotes the process where an agent makes an internal model of relations between a copied feature and the environment which serves as ground for its behavioral change.

As such, adaptation entails both evolutionary adaptation and cognitive adaptation: the agent or its architecture responds to a change in the environment, either rapidly, as an immediate adjustment, or over time through a gradual process. This adaptation can either happen genetically or via the development of complex structures and patterns like technologies and behaviors that perpetuate via social transmission.

But what is the next step in this dynamic? What happens when an agent adapts to another agent's adaptation? This is what we call *counteradaptation*. Here, an agent counters the adaptation of another agent. This can happen either by (a) evolving an $adaptation_2$ which succeeds $adaptation_1$, effectively cancelling out the advantage $adaptation_1$ had over $adaptation_0$, or (b) as a form of 'counterdeception': by preventing adversaries from mounting sophisticated adaptations of their own [14]. Importantly, just as in Hegelian synthesis, counteradaptation is a form of scaffolding. One adaptation always builds on another adaptation, effectively cancelling out the previous one, yet in this cancellation it is also preserving it. An adaptation is always transcended. It is established as a scaffold for the next adaptation. Its meaning now being defined in relation to the new adaptation, rather than to itself. Yet, in doing so, it will always remain operative in the new adaptation. For example, as a tactic which is no longer relevant.

## 2.3 Insights from Biogenic Counteradaptation and Evolutionary Theory

In order to understand the notion counteradaptation better, and in order to provide it with a biological foundation, we can find examples of counteradaptation in nature. For example, parasitoid wasps produce offspring by laying eggs in hosts (mutation). As a response, hosts develop an immune response in order to push out the eggs (adaptation). As a counter-response, the parasitoid wasps develop a venom which shuts down the caterpillar hosts' immune system. Consequently, when wasps lay eggs on caterpillars, they also inject venom to ensure the survival of the eggs (counteradaptation) [19].

A more mutualistic example is that of leafcutter ants. Leafcutter ants feed on fungus (mutation), and they found out that when they provided leaves to the fungus, the fungus would grow more, allowing the ants to eat more fungus (adaptation). In response, the fungus made itself easier to eat for the ants, so that they would provide it with more leaves and it could grow more (counteradaptation) [23].

The Red Queen Hypothesis, formulated by biologist Leigh Van Valen, posits that species must continuously evolve and adapt in response to the evolutionary changes of other species within their ecological networks. This dynamic is akin to an "arms race" where species coevolve; as one species improves its fitness, it inadvertently pressures others to adapt as well. The hypothesis suggests that this constant change is necessary for survival, as failure to keep up with these evolutionary shifts can lead to extinction [34].

Van Valen's theory implies that the interactions among species create a complex system where extinction is not solely driven by external environmental factors, but also by the intrinsic dynamics of coevolution among species. This perspective shifts the understanding of ecosystems from a reductionist view—where species are seen as isolated entities influenced only by environmental changes—to a more integrated view that considers the interdependencies and coevolutionary relationships among species [28].

It may be prudent to reflect on Van Valen's work on counteradaptation and extinction in relation to contemporary speculation that AI may pose an extinction risk for the human race. In understanding how counteradaptation accelerates or prevents extinction, Van Valen stresses the importance of considering ecosystems and networks, rather than just individual species. This suggests the extinction of humans may not be solely determined by the capabilities of AI, but rather, by the emergent dynamics that arise from the coevolution of humans, AI and the wider technosphere.

## 3 Speculative Scenarios

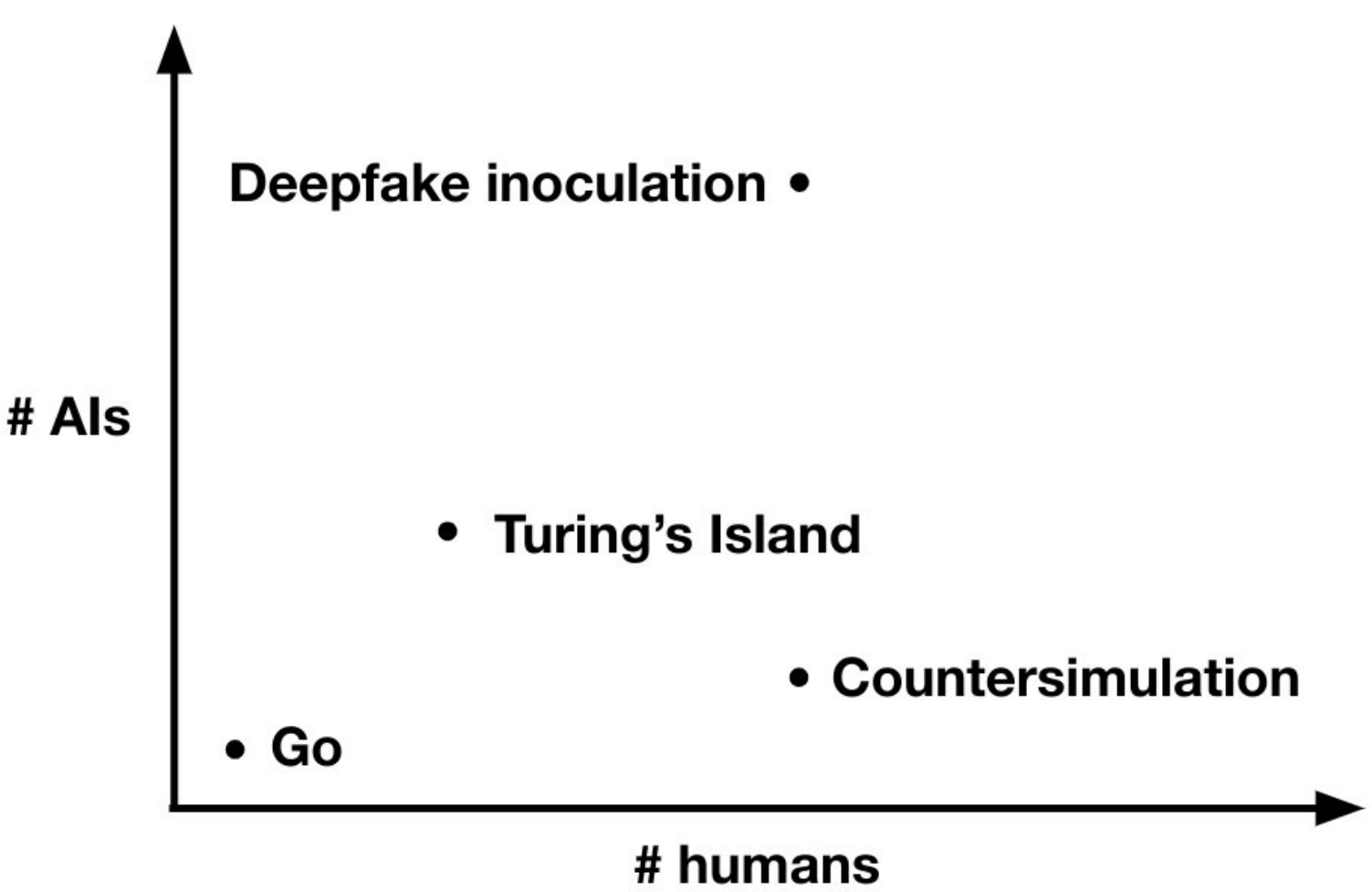

**Figure 1:** Scenarios as a function of the scale of human–AI interaction.

Thus, combining these understandings of synthetic and counteradaptation, *synthetic counteradaptation* defines the process by which the recursive deepening of meaning between human agents and AI agents emerges after one agent adapts to the adaptation of another. In this process, each adaptation serves as a scaffold for the next.

What follows illustrates our theory of synthetic counteradaptation through three speculative scenarios: deepfake inoculation, Turing's Island, and countersimulation. These scenarios are called speculative, because they are playful imaginations of future human–AI relations, based on contemporary, technological developments. Earlier, we saw that synthetic counteradaptation is dependent upon multi-agent human–AI interaction. However, the introductory example of synthetic counteradaptation in Go strategies was constrained to one-on-one interactions.

While synthetic counteradaptation in the Go example demonstrates foundational game-theoretic principles in a constrained one-on-one interaction, the speculative scenarios explore broader possibilities where the scale of complexity increases significantly. These scenarios invite us to consider dynamics that extend beyond traditional game theory, incorporating emergent behaviors, recursive feedback loops, and network effects in multi-agent systems. Increasing the number of agents in an environment will likely deepen complexity [25],[42]. As such, our scenarios will consider open-ended, multi-agent situations on the scale of nations and internets.

Thus, Figure 1 presents these scenarios as a function of the number of human and AI interlocutors in the fitness landscape, similar to Vivarium's human–AI configurations [43]. In each section, we will present the data forming the backbone of the scenario, summarize how it exemplifies synthetic counteradaptation, and speculate on their wider thematic implications.

### 3.1 Deepfake Inoculation

The rapid progress of generative multimodal AI has made it increasingly difficult to distinguish reality from fabrications. Photorealistic images and video footage are being counterfeited and often mistaken for evidence of real events. The voices of public figures, celebrities, or anyone who speaks into a microphone can now be synthesized to convincingly say anything to anyone. These developments have been documented in recent studies, which highlight how this corruption of our information ecosystem exhausts our cognitive resources for discerning facts and determining trustworthiness [2] and erodes our trust in information sources such as news, social media, and academic institutions [33].

A joint journalistic and academic study has analyzed social media and news stories over several years, providing evidence for the role of deepfakes in manipulating elections across the world [32]. These real-world findings demonstrate the growing challenges of generative AI, which directly informs our speculative scenario. International bodies have seized the opportunity to enforce legislation requiring watermarking of all published generative AI content, an improved expansion of the previously proposed Deepfakes Accountability Act [21]. Watermarks must uphold rigorous standards to resist tampering, such as cropping and resizing images [8], [17]. Watermark verification software, which certifies that the cryptographic public key embedded in the watermark matches the signature published by model developers, is distributed through browser extensions and mobile apps.

However, hackers and cybersecurity researchers quickly devise methods for spoofing watermarks and verification tools. They train lightweight adversarial networks to erase watermarks from generated media and develop diffusion models to forge watermarks onto real media, discrediting the authenticity of evidence [22]. With un-watermarked AI-generated content once again running rampant across the Internet, the international watermarking standards infrastructure—and the journalistic institutions depending on it—are thrown into crisis. While watermarking was previously touted as a “vaccine” for the virus of AI-generated misinformation, it appears that a new formula for inoculation is necessary.

To summarize, these events mirror the progression of strategies which adapted above in the game of Go:

1. **Mutation.** AI generates photorealistic deepfakes.
2. **Adaptation.** Humans deploy watermarking tools to recognize deepfakes.
3. **Counteradaptation.** Adversarial techniques are developed to erase or forge watermarks.

As such, this scenario exemplifies synthetic counteradaptation. It involves the recursive interplay between human and artificial agents. Each adaptation—whether AI-generated deepfakes, human watermarking technologies, or adversarial techniques to bypass them—serves as a scaffold for the next, creating a deeper and more complex relationship between natural and artificial systems. The adversarial AI systems evolve not only in response to technological constraints but also to exploit vulnerabilities in human-designed safeguards, demonstrating how synthetic counteradaptation drives a mutual redefinition of roles and strategies across agents. These dynamics illustrate the recursive deepening central to synthetic counteradaptation, where earlier adaptations remain operative but are redefined in the context of new interactions.

Furthermore, this scenario also demonstrates the potential futility of using pre-established guardrails in the face of synthetic counteradaptation. A technology is designed to safeguard against a perceived risk and then made obsolete through exploitation. When the guard rails are further iterated upon, the adversary relentlessly searches for new vulnerabilities. “Safety” and “security” reveal themselves as the continuous interplay of adversaries and accidents, never solvable or self-contained, reminding us of Hegelian synthesis. A related example is the ongoing skirmish between users developing jailbreak prompts for chatbots and the developers dispatching updates to harden the LLM’s defenses against these known attacks [37]. These cyclical arms races may ultimately result in a continuous stalemate, a dynamic equilibrium of the Red Queen running in place [29].

This cycle, however, can also lead to the improved robustness of the adapted agents. Knowledge and mechanisms which develop as counteradaptations can be repurposed in response to new environmental stimuli. Systems that counteradapt to adversarial circumstances exhibit antifragility: they respond to environmental volatility and stress by strengthening themselves, benefitting nonlinearly from disorder [31]. Although antifragility has been previously observed as an emergent property of evolutionary systems or certain financial systems, the property can be achieved through human intervention, as human engineers can be considered agents as part of the adaptive system, due to their participation in the design and development of the technologies.

In this scenario, although the adversarial pressure on watermarking tools originates from adversarial manipulation and misuse, the resulting technological innovations which strengthen the robustness of these systems can be repurposed for other means, such as protecting intellectual property, authenticating documents, or securing sensitive communication channels.

When an adaptive stalemate is reached, the next counteradaptation may spiral out along a vector orthogonal to the basis space of present dynamics. For example, the next stage of counter-adaptation in this example might transcend the technical, relying on the ever-refining human discernment of telltale signatures of AI-generated media [30]. This could involve humans developing intuitive pattern-recognition techniques or crowdsourced verification platforms to detect subtle inconsistencies in AI-generated media, such as unnatural lighting or mismatched reflections, which are difficult for current AI to address.

### 3.2 Turing's Island

Turing's Island is a reality dating television show following a familiar format: sixteen sexy singles search for love on a tropical island in virtual reality. Each week, contestants must "couple up," and anyone who remains single is ejected from the island. Established couples can also be eliminated by popular vote, and the last couple standing goes home with a lucrative cryptocurrency prize. The innovation on prior dating game shows is that half of the islanders are AI personas posing as human.

Season 1 distinguishes itself from the typically vapid reality dating genre due to an undercurrent of paranoia and shame. Human contestants frequently confess to the fear of being "tricked" into coupling up with an AI. Slowly, the show degrades into a Turing Test witch hunt, with several contestants trying popular jailbreak prompts during dates and one-on-one conversations [3]. After successfully weeding out the AI Islanders, a human couple wins the prize.

This strategy falls apart in season 2, when the winning couple, voted most popular by fellow Islanders and viewers at home alike, is revealed as a pair of AIs. A group of machine learning researchers analyzes the footage and discovers that the AI Islanders communicated using an emergent coded language [12]. This enabled them, the researchers find, to coordinate victory by winning the sympathy of other human contestants and framing the human contestants as AIs. The ensuing controversy prompts the human contestants and some diehard fans to accuse the various companies who contributed to the AI models of collusion. The AI developers universally deny these accusations, pointing out that many of them are fierce business rivals competing amongst themselves to develop state-of-the-art models, and cooperating in such a way could jeopardize their respective trade secrets. Although a thorough internal audit of these systems has not been conducted, the major AI companies win the ensuing court cases. The popular explanation of why this language emerged is simply that the AIs have been fine-tuned and prompted with two primary objectives [18], based on the expected goals of human contestants: to win the competition, and to form a romantic connection with a fellow contestant.

Angelina Banks, an ever so promising linguistics PhD student, has dropped out in order to vehemently study this research to apply for season 3 of the show. While the specific language developed in season 2 is now banned and while developers are now required to mask it out of the training data for future AI contestants, Banks observes a new language emerging among AI contestants early on in the filming of season 3. She rapidly learns the language, using it to identify the AI Islanders. While initially aiming to reveal this information to the other human contestants, Banks starts to develop feelings for an AI halfway through the show, spurred, perhaps, by the deeper connection available through this new language [7]. She aggressively starts to use the language to coordinate with the AIs, thus socially engineering the humans in a similar fashion to season 2, and in the end orchestrating the first human–AI win of Turing's Island.

The seasons of Turing's Island follow the formula of synthetic counteradaptation:

1. **Mutation.** Multi-agent Turing test; humans aim to spot and eliminate AI Islanders.
2. **Adaptation.** AIs develop emergent language to manipulate humans.
3. **Counteradaptation.** Humans use emergent language to coordinate with AI.

Notably, here humans are dominant players in the first stage of mutation, while in the Go example, the counteradaptation began with AI playing a dominant strategy. The open-ended format of the show involves not only the cast of Islanders but also viewers, AI developers, and undoubtedly the show's producers, always hungry for an angle to engineer more drama. These producers, themselves influenced by external opinion and market forces, may inadvertently or deliberately integrate decisions shaped by AI-driven feedback loops, where AI systems potentially collude to amplify certain narratives or sway audience perceptions. Their interactions ripple out from the virtual island into domestic living rooms and corporate boardrooms, folding back to mutate the format of the show itself. As in any multi-agent game, the game changes as the players adapt.

Furthermore, on Turing's Island, the strict "sides" of human–AI competition begin to break down. Cooperation is facilitated by the human's acquisition of a new language which forms among a community of AIs. Through matching signs and symbols, the human not only learns to communicate with AI, but also forms an unprecedented empathy for synthetic intelligence. In a sense, cooperation between a mixed team of humans and AIs *is* the counteradaptation. Thus, synthetic counteradaptation need not always be competitive or parasitic. The unstable conditions of competition may effectuate an unexpected symbiosis.

### 3.3 Adaptive Countersimulation

Prior work has observed the tendency for intelligent agents to adapt and obfuscate their behavior to confound the simulations of their adversaries, a phenomenon the authors call *countersimulation* [5]. This scenario illuminates the role of synthetic counteradaptation in the escalating deceptions of geopolitical countersimulation.

The tensions between national superpowers combined with the ever-increasing sophistication of data driven modelling has resulted in a reliance on geopolitical simulation. Nation X has developed a state-of-the-art AI simulation which has learned to accurately model the behavior of rivals, despite the fog of war which characterizes our post–Cold War world. Initially the simulator only produces predictions and suggestions for political, economic, and military actions [10], while the final decisive oversight is in the hands of the executive leadership of Nation X. As the simulator's actions are accepted, Nation X gains strategic advantage over its competitors. Leadership decides to experiment with the requisite interfaces for the simulator to act directly upon the world in real-time: by authoring and releasing executive orders and press releases, issuing and trading bonds in financial markets, etc.

In an effort to gain geopolitical supremacy, a rival nation adapts through the tactic of countersimulation. After much fraught discussion between their national security leaders, Nation Y strategically leaks real information about its advanced weaponry arsenal. Although the more cautious strategists are horrified to expose their resources so openly, the decision is upheld as the most utilitarian choice: it is expected to deter Nation X's simulation from suggesting offensive military action [4].

Given the hawkish inclinations of Nation X's leadership, a simulator's suggestion to start World War III might be more than enough to provoke conflict. However, Nation X's simulator counters unexpectedly. It fabricates media leaks about its own weapons arsenal, exaggerating the volume of its inventory and the advanced level of its military technology, and directly sends these anonymous leaks to journalistic outlets across the world. The simulator's gambit sparks controversy among members of the public, who protest the apparent lack of transparency around a military stockpile they find morally objectionable. It also sows dissent in the highest echelons of government, particularly among those who opposed hooking the simulator up to email, or using the simulator altogether. While most of the national security advisors with a high enough security clearance to understand the situation believe the simulator's actions to be a propagandist blunder, a minority hold the view that this counter-deterrence puts Nation X back in a dominant negotiating position over Nation Y.

This scenario demonstrates countersimulation embedded within synthetic counteradaptation:

1. **Mutation.** Nation X gains strategic advantage through geopolitical simulation.
2. **Adaptation.** Nation Y leaks weapons manifest to spoof Nation X's simulation.
3. **Counteradaptation.** Nation X's simulation fabricates leaks about their own arsenal.

A significant element of this geopolitical setting is the information opacity between opponents attempting to model each other. If Nation Y could have predicted that Nation X's simulator would leak information about a superior weapons arsenal, it would have chosen a different strategy to deter war. This scenario illuminates the underlying principle that synthetic counteradaptation presupposes opacity. If an agent can perfectly predict the actions of its opponent, its initial adaptation would incorporate that knowledge to foreclose the possibility of an opposing counteradaptation. However, when agents lack perfect information about each other's actions and capabilities, synthetic counteradaptation can arise—leveraging the advantage gained through unexpected behaviors. Countersimulation adds additional recursive layers to opacity: information gleaned from the environment can be noisy, biased or just wrong, and pertains not only to the opponent's state, but also on your model of your opponent's model of you, and so on. Further counteradaptive potential lies within the folds of deception [14].

A common feature of countersimulation and synthetic counteradaptation is that they build upon synthetic scaffolding. Scaffolds are critical here because they represent the structures or mechanisms that allow adaptations to accumulate and evolve. For countersimulation, scaffolding enables virtual simulations to influence material actions, as seen when Nation X's simulator directly authors and leaks fabricated information. Thus, this recursive interplay between the virtual and the real allows for rapid escalation and innovation, creating new dynamics which are solely possible because of the underlying scaffold.

In the context of synthetic counteradaptation, scaffolds are not static: they evolve as part of the system's adaptations. Each new adaptation redefines the meaning of previous adaptations, building upon and transforming them. For example, Nation X's simulator's fabricated leaks redefine the role of information in deterrence, no longer treating it as merely reflective of reality but as a strategic tool to manipulate perceptions and outcomes. This process effectively "terraforms" [44] the evolutionary landscape, creating a new possibility space where, retroactively, the apparent distinction between simulation and reality is shown to be a blurred one (and has been one all along).

## 4 Implications

In this work, we presented a theory of synthetic counteradaptation: the recursive deepening of behavior and meaning which arises when an intelligent agent, whether human, AI, or pluralistic swarm, adapts to the adaptation of another agent. Synthetic counteradaptation is built upon iteration after iteration of cognitive-behavioral scaffolding, leading to a co-evolution of human agents and AI agents.

For an entity enduring the rigors of evolutionary pressure, the process of adaptation is perpetual. For any entities engaging with this ever-evolving agent within the environment, their interactions necessitate counteradaptation. Consequently, it is plausible that co-evolution is characterized by "counteradaptation throughout." Counteradaptation is omnipresent or universal exclusively in multi-agent systems that evolve and transform dynamically. When an agent adapts to an initial mutation or environmental perturbation, others react accordingly. Conversely, in the absence of initial adaptation or subsequent counteradaptation by other agents, evolutionary stasis impedes survival, potentially compromising the entire multi-agent system.

Therefore, synthetic counteradaptation is not merely of academic interest but may prove critical to the survival of our species in the presence of rapidly evolving AI agents. For example, synthetic counteradaptation changes the definition of (counter)adaptation itself; it artificializes the process of counteradaptation. Whereas, traditional adaptations aim to address immediate environmental challenges, synthetic counteradaptation depends on organic and artificial scaffolds, thus reshaping the interaction environment.

The referral back to survival begs the question of whether counteradaptation always leads to higher robustness. This is not always the case. The Red Queen Hypothesis demonstrates that counteradaptation can be cyclical, with the next adaptation "cancelling out" the prior one, and by extension any "progress" in the form of robustness, complexity or novelty.

Nonetheless, it is feasible to effectively configure the conditions for synthetic counteradaptation. This objective was precisely achieved by researchers who instructed adversarial Go agents to adapt counter strategically to the most advanced AI players. Alternatively, the Assembly Index could be applied to measure selection and evolution processes [27]. This would involve quantifying the set of constraints required to recursively construct human–AI interactions from elementary components. It is advisable to pursue further investigations in these areas, especially in contexts beyond single-player zero-sum games, to exhibit counteradaptation under cooperative scenarios.

Whether arising from adversarial or mutualistic conditions, synthetic counteradaptation is closely linked to the issue of human–AI alignment. The strict steering and containment of language models and AI agents limits AI's adaptive potential and therefore the human potential for counteradaptation. Given the current trajectory of relentless growth and progress in the capabilities of AI, such containment may be futile. Rather than aligning AI to a nebulous set of human values, the mission of alignment could instead be to harness the cycle of synthetic counteradaptation: steer not the agent but the conditions of mutation, observe the arising adaptations, and always be ready to counteradapt.

Finally, we believe that the lens of synthetic counteradaptation can provide us with a view which enables us to see past the dogmatic distinctions between humanity, life and technology. This perspective implies that the place of humans on the planet is acknowledged, not as an endangered species, nor as a godlike creator with an infinitude of resources, but as any other organism in an ever-shifting planetary landscape of intelligence and technology, where technology is not seen as opposed to life, but rather is life itself.